\documentclass[a4paper,man,natbib,floatsintext]{apa6}

\usepackage[english]{babel}
\usepackage[utf8x]{inputenc}
\usepackage{amsmath}
\usepackage{graphicx}
\usepackage[colorinlistoftodos]{todonotes}

\usepackage{xcolor}
\usepackage{algorithm}
\usepackage{algorithmic}
\usepackage{subcaption}
\usepackage{graphicx}
\usepackage{multirow}
\usepackage{booktabs}
\usepackage{lipsum}
\usepackage{tcolorbox}
\usepackage{multicol}
\usepackage{dcolumn}
\usepackage{array,ragged2e}
\usepackage{setspace}
\usepackage{float}
\usepackage{endnotes}
\let\footnote=\endnote

\usepackage{url}

\usepackage{breakurl}
\usepackage[breaklinks]{hyperref}

\title{The ``Collections as ML Data'' Checklist for Machine Learning \& Cultural Heritage}
\shorttitle{The ``Collections as ML Data'' Checklist}
\author{Benjamin Charles Germain Lee}
\affiliation{
University of Washington\\
\href{mailto:bcgl@cs.washington.edu}{bcgl@cs.washington.edu}
}

\abstract{
    Within the cultural heritage sector, there has been a growing and concerted effort to consider a critical sociotechnical lens when applying machine learning techniques to digital collections. Though the cultural heritage community has collectively developed an emerging body of work detailing responsible operations for machine learning in libraries and other cultural heritage institutions at the organizational level, there remains a paucity of guidelines created specifically for practitioners embarking on machine learning projects. The manifold stakes and sensitivities involved in applying machine learning to cultural heritage underscore the importance of developing such guidelines. This paper contributes to this need by formulating a detailed checklist with guiding questions and practices that can be employed while developing a machine learning project that utilizes cultural heritage data. I call the resulting checklist the ``Collections as ML Data'' checklist, which, when completed, can be published with the deliverables of the project. By surveying existing projects, including my own project, Newspaper Navigator, I justify the ``Collections as ML Data'' checklist and demonstrate how the formulated guiding questions can be employed and operationalized. 
}

\begin{document}
\maketitle

\section{Introduction}

Over the past few years, the field of machine learning has seen a growing movement to develop algorithmic impact assessments, checklists, and best practices that practitioners can consult while creating datasets, training models, and operationalizing these systems \citep{ data_statements_NLP,mitchell_model_2019}. These efforts show great promise, having seen widespread adoption across the field -- from conference submission requirements \citep{NAACL_reproducibility_checklist,Pineau2021ImprovingRI} to researchers' utilization of these resources to guide their own work and communicate their decisions throughout a project's development.

Concurrently, practitioners within the cultural heritage sector are critically investigating the sociotechnical implications of applying machine learning to digital collections \citep{alpert-abrams_machine_2016,cordell_q_2017,cordell_machine_nodate,eventsummaryLCLabs,lee_compounded_2020,UNL_report,padilla_responsible_2020,europeanaInterimReport}. Indeed, machine learning has a long history among the gallery, library, archive, and museum (``GLAM'') communities, as well as the digital humanities: for example, optical character recognition (OCR) algorithms have been used in digitization pipelines for decades \citep{cordell_machine_nodate}. Informed by critical cataloging \citep{adler_classification,bowker_sorting_2000,furner_dewey_2007,knowlton_three,olson2001power}, critical data studies \citep{excavating_ai}, science \& technology studies \citep{noble_algorithms_2018}, the ``Collections as Data'' movement \citep{padilla_2018, collections_as_data}, best practices and digital strategies among cultural heritage institutions \citep{lcdigitalstrategy}; and decades of scholarship by librarians and archivists, this effort has produced ``state of the field'' reports \citep{cordell_machine_nodate} and best practices for cultural heritage institutions \citep{padilla_responsible_2020}. Reports such as Thomas Padilla's ``Responsible Operations: Data Science, Machine Learning,
and AI in Libraries'' have been foundational in articulating principles and provocations to guide the operationalization of machine learning at cultural heritage institutions \citep{padilla_2018}. Ryan Cordell's ``25 Questions for Structuring an ML Project'' builds on this work to provide a set of guiding questions for specific projects involving machine learning and libraries, a crucial development \citep{cordell_machine_nodate}.

Consequently, there is much to be gained by further developing checklists specifically for practitioners carrying out machine learning projects in the context of treating cultural heritage collections as data. Practitioners engaging in projects in this liminal space must be thoughtful in every step of the the project's development, including framing, implementation, release, and maintenance. Even with good intentions, practitioners embarking on projects in this space risk kitschifying or exploiting those represented in the digitized collections in question; glossing over digitization subtleties that impact the performance and output of machine learning models; utilizing machine learning when it is not necessary due to organizational agendas surrounding emerging technologies; beginning machine learning projects with no plans for sustainability; or violating the privacy of end-users of systems that are built. Proper usage of a checklist facilitates thoughtful engagement with such challenges and therefore has the potential for profound impacts for practitioners and institutions. 

This paper draws from these emerging movements in the machine learning and cultural heritage communities in order to produce the ``Collections as ML Data'' checklist that researchers and practitioners alike can utilize when embarking on projects in machine learning and cultural heritage. The completed checklist can be published with a project's deliverables in order to provide the audience with a structured description of the project's considerations and limitations.

\section{Related Work}\label{sec:related_work}
This paper contributes the first detailed checklist for projects at the intersection of machine learning and cultural heritage. The ``Collections as ML Data'' checklist builds on existing work surrounding best practices within the respective machine learning and cultural heritage communities, as well as the growing scholarship at the intersection of these fields. It also draws from the literature on conceptual models and scholarship on the digital humanities, critical data studies, and science \& technology studies within the broader umbrella of information studies. This section serves to contextualize this related work in more detail.

\subsection{Checklists, Toolkits, and Best Practices within the Machine Learning Community}\label{sec:mlliterature}

This paper draws from existing checklists, toolkits, and best practices surrounding machine learning. Relevant work includes  \citep{ainow, katell_algorithmic_2019, madaio_co-designing_2020, mitchell_model_2019, gebru2020datasheets, holland2018dataset, data_statements_NLP, data_cards, data_catalogs, arnold2019factsheets, ribeiro_beyond_2020, consumer_labels, system_cards, lacoste2019quantifying, friedman_envisioning_2012, tarot_cards, diakopoulos_algorithmic_2014, diakopoulos_2016}.\footnote{It is worth noting that domain-specific machine learning checklists and best practices have been published as well, for example with chemistry \citep{chemistry_checklist} and medicine \citep{clinician_checklist}.} In the section ``\nameref{sec:results1},'' a more detailed taxonomy of these guidelines is provided.

Collectively, these checklists, toolkits, and impact assessments emphasize a range of considerations for different stages of machine learning projects, including the construction of a dataset, the training and auditing of a model, and the deployment of a model in an operational sense. In addition, the checklists target a range of different audiences, from researchers and practitioners themselves to downstream users of the constructed datasets or trained models. As highlighted in \citep{data_statements_NLP}, such considerations are significant because ``there are both scientific and ethical reasons to be concerned. Scientifically, there is the issue of
generalizability of results; ethically, the potential
for significant real-world harms.'' Indeed, from a scientific standpoint, incomplete documentation of a dataset or model can lead to unintentional misuse by other practitioners; insufficient evaluation of a dataset or model can lead to unforeseen issues of generalization when operationalized; and lack of clarity surrounding copyright can hinder the adoption of a dataset or model. From an ethical standpoint, machine learning datasets risk exploiting personal data and raising questions of privacy; labor practices behind data annotations are not always foregrounded, leading to questions of labor exploitation; machine learning models can perpetuate marginalization and oppression; and operationalized systems can be fragile, failing without warning to end-users. Though no set of guidelines can comprehensively cover all such scientific and ethical questions, a checklist nonetheless represents a first significant step toward systematizing shared practices surrounding ethical and responsible decision making. In this regard, the checklist represents a conceptual model for analysis with emerging machine learning methodologies. While all of the aforementioned checklists apply to the machine learning community, it is worth noting that none of them have been designed with the setting of cultural heritage particularly in mind -- precisely the setting that the ``Collections as ML Data'' checklist contributed in this paper adopts.

\subsection{Best Practices from Cultural Heritage}\label{sec:ch}
This paper draws from an emerging body of literature devoted to best practices surrounding machine learning and data science within cultural heritage. Relevant literature includes the Library of Congress Labs team’s summary report for the ML + Libraries summit  \citep{eventsummaryLCLabs}, Ryan Cordell’s report ``Machine Learning + Libraries: A Report on the State of the Field'' \citep{cordell_machine_nodate}, Thomas Padilla’s OCLC report ``Responsible Operations: Data Science, Machine Learning, and AI in Libraries'' \citep{padilla_responsible_2020}, and the Europeana ``AI in Relation to GLAMs'' Task Force's 2021 report \citep{europeanaFullReport}. Collectively, these reports draw from a long history of critical work within library, archival, and information sciences, as well as the humanities, in order to articulate the specific subtleties that arise when applying emerging technologies to cultural heritage in particular: how to uphold the library community's standards for privacy, how to consider the provenance of collections and their digitization, and how to foster technical fluency among cultural heritage practitioners are just a few such considerations \citep{padilla_responsible_2020}. This body of work in cultural heritage thus proposes conceptual models for ethical practices when applying new technologies to digital collections.

While these reports foreground best practices and recommendations, they are intended primarily as guiding principles at the organizational and field level, rather than at the level of a machine learning practitioner embarking on a specific ML project in cultural heritage. Only Ryan Cordell’s report offers a checklist for practitioners; the checklist takes the form of ``25 Questions for Structuring an ML Project'' \citep{cordell_machine_nodate}. The ``Collections as ML Data'' checklist proposed in this paper builds on this work to produce a detailed checklist that, when completed by practitioners, can be distributed with project deliverables in order to convey the considerations and limitations surrounding the project to its audience and auditors.

\subsection{Machine Learning, Cultural Heritage, Information Science, and the Digital Humanities}\label{sec:surveys}

This paper builds on the rich collective body of projects and research at the intersection of machine learning and cultural heritage. Surveys and state of the field reports such as \citep{fiorucci_2020,cordell_machine_nodate,europeanaFullReport,reviews_in_dh} describe the history of such work, as well as enumerate current projects. 
In order to develop, refine, and evaluate the proposed checklist with case studies, this paper draws from these surveys and state of the field reports at the intersection of machine learning and cultural heritage. The chosen projects are described in more detail in the section ``\nameref{sec:results2};'' here, I note that I utilize my project, \href{https://news-navigator.labs.loc.gov}{Newspaper Navigator}, as a case study because experiences developing the dataset \citep{nn_dataset} and launching the search application \citep{nn_demo} served as initial provocations for developing this paper. For more information on the autoethnographic approach that I adopted for Newspaper Navigator, I refer the reader to \citep{lee_compounded_2020}.

More generally, it is worth situating the ``Collections as ML Data'' checklist within the broader landscape of research within information science and the digital humanities, both of which have historically cultivated research at the intersection of machine learning and cultural heritage \citep{JASIST_special_issue}. Both disciplines have embraced novel information artifacts as  worthy of publication alongside traditional journal papers. Such artifacts include datasets,\footnote{See, for example, the \textit{Journal of Open Humanities Data}.} online visualizations and exhibitions, and computational replay \&  provenance systems. These efforts represent innovative deliverables that foreground transparency, making it possible for research communities to more easily reconstruct research findings as well as pursue new research on top of existing scaffolding. The ``Collections as ML Data'' checklist aligns squarely with this pursuit of transparency within information science and the digital humanities by serving as a deliverable that, when completed and published, addresses precisely these goals. In this regard, the ``Collections as ML Data'' checklist serves as a conceptual model of the sociotechnical considerations undertaken by a researcher or practitioner applying machine learning to cultural heritage.
\section{Overview of Methodology}\label{sec:methodology}

In order to formulate initial questions and suggestions for the ``Collections as ML Data'' checklist from a machine learning perspective, I started by surveying the relevant machine learning literature. To make sense of this capacious space of checklists, toolkits, impact assessments, and beyond, I produced a taxonomy of such work. One goal of creating this taxonomy was to develop a more comprehensive list of guiding questions to be included in the ``Collection as ML Data'' checklist from the perspective of machine learning. A second goal of producing this taxonomy was to help situate cultural heritage practitioners with a guiding roadmap for existing work from the machine learning literature. Accordingly, I present this taxonomy later in this paper.

Next, I utilized the reports and surveys on machine learning projects in cultural heritage in order to select a representative grouping of projects as case studies for developing and testing the ``Collections as ML Data'' checklist. In particular, I refined desiderata for selecting projects that emphasized project diversity: institutional setting, digitized collection medium, digitized collection topic, machine learning methodology employed, field of study, intended audience, and final deliverable form. This process yielded five projects to serve as case studies. Though many papers in the machine learning checklist literature utilize case studies to evaluate the proposed checklists, the methodology behind selecting the case study projects is not always detailed. I elaborate on this process in ``\nameref{sec:results2}'' with the goal of transparent documentation.

The next step entailed surveying best practices and responsible operations from cultural heritage in order to develop checklist questions specific to cultural heritage projects. Here, I primarily drew from Ryan Cordell’s report \citep{cordell_machine_nodate}, Thomas Padilla’s report \citep{padilla_responsible_2020}, LC Labs’s ML + Libraries summit report \citep{eventsummaryLCLabs}, and the ``EuropeanaTech AI in relation to GLAMs'' Task Force’s report \citep{europeanaInterimReport}. As described in the section entitled, ``\nameref{sec:ch},'' these papers contain various guidelines and principles for machine learning and cultural heritage, touching on the special sensitivities to be considered by cultural heritage practitioners, as well as the challenges faced by collaborations among stakeholders with different levels of fluency and training with machine learning and cultural heritage. I translated these guidelines and provocations into questions within the ``Collections as ML Data'' checklist.

I next turned to refining the checklist utilizing existing projects in this space. Drawing from my own experiences with my  project, Newspaper Navigator \citep{lee_compounded_2020,nn_demo,nn_dataset}, which served as the initial motivation for developing this checklist, I formulated a series of additional questions relevant to working with cultural heritage collections, resulting in an initial version of the ``Collections as ML Data'' checklist. To test this initial version, I utilized the projects selected as case studies. Inspired by the methodology utilized in \citep{critical_race_theory}, these case studies resulted in vignettes for justifying and detailing each checklist item. In analyzing each project, I identified redundancies within the checklist questions as well as subtleties raised by the project that were not yet covered by the checklist. Accordingly, I iteratively revised and refined the checklist with each case study.

The last step in refining the checklist entailed incorporating feedback from practitioners and researchers in both machine learning and cultural heritage. I further tested the efficacy and extensibility of the checklist by workshopping it with colleagues at the University of Washington and other institutions. I began by soliciting feedback from Professor Katharina Reinecke and the CSE 599 ``Computer Ethics'' graduate course at the University of Washington. I subsequently solicited feedback from researchers and practitioners specializing in cultural heritage. This step culminated in the ``Collections as ML Data'' checklist.
\section{A Taxonomy of ML \& AI Toolkits, Checklists, and Impact Assessments}\label{sec:results1}
As described in the section entitled, ``\nameref{sec:methodology},'' in order to develop initial questions for the ``Collections as ML Data'' checklist based on existing work, I performed a literature review for checklists, impact assessments, toolkits, and best practices in machine learning. I began the review with papers that I had already encountered and followed the citation graphs. In addition, I searched paper repositories and consulted colleagues within machine learning. Lastly, I sought recommendations from colleagues in machine learning. With the literature identified, I then categorized papers into four discrete groups. The collected works in each of these four categories offer a different perspective for machine learning projects, and one explicit aim of the ``Collections as ML Data'' checklist is to draw from all of these perspectives. It should be noted that the set of papers that I have included is not comprehensive, as work in this space is evolving quite quickly, and I limited my survey primarily to academic publications. As such, this taxonomy should not be treated as a full survey but rather a guiding resource for cultural heritage practitioners looking to situate themselves within the landscape of machine learning assessments.

\subsection{Dataset Assessments}\label{sec:data}

Though datasets have always been essential to the field of machine learning, it has been only in recent years that the discipline has begun to adopt a critical lens toward examining their construction, composition, and utilization. Often motivated in the literature through invocations of examples of machine learning models being deployed in high stakes decision-making processes (i.e., medical diagnosis, legal recidivism, and credit score determination), an emerging body of work is calling for practitioners to publish dataset assessments along with the datasets themselves. The dataset assessment can serve as a compliance checklist for practitioners (such as ensuring IRB \& GDPR compliance \citep{goodman_european_2017}); a practical guide for consumers of the dataset (such as other machine learning practitioners using the dataset to train a model); a pedagogical tool for the public; and a means for recourse for those whose data are contained within the dataset. Data assessments are often inspired by and modeled after regulatory efforts such as the nutrition label for food and beverage packaging. Five related approaches to data assessments that have received widespread attention in the machine learning community to date are:
\begin{enumerate}
    \itemsep0em 
    \item ``Datasheets for Datasets'' \citep{gebru2020datasheets}
    \item ``Data Statements for NLP'' \citep{data_statements_NLP}
    \item ``The Dataset Nutrition Label'' \citep{holland2018dataset}
    \item ``Data Cards'' \citep{data_cards}
    \item ``Comprehensive and Comprehensible Data Catalogs'' \citep{data_catalogs} 
\end{enumerate}

\subsection{Model Assessments}\label{sec:model}

A concurrent thread emerging from the machine learning literature is the development of model assessment rubrics and checklists for practitioners to complete. Motivated in a similar fashion to dataset assessments, model assessments concern the analysis of the training, evaluation, deployment, and operationalization of the model itself. Existing work in this space has advocated for these model assessments to be published along with the models themselves, much akin to dataset assessments. Examples of model assessments include:

\begin{enumerate}
    \itemsep0em 
    \item ``Model Cards for Model Reporting'' \citep{mitchell_model_2019}\footnote{Closely related work, such as ``Interactive Model Cards'' \citep{interactive_model_cards}, is worth noting as well.}
    \item ``CheckList for NLP Models'' \citep{ribeiro_beyond_2020}
    \item ``FactSheets'' \citep{arnold2019factsheets}
    \item ``Consumer Labels'' for machine learning models \citep{consumer_labels}
    \item ``System Cards'' for AI decision-making for public policy \citep{system_cards}
    \item Microsoft Research's AI fairness checklist \citep{bird2020fairlearn}
    \item The ``AI Blindspot'' discovery process \citep{ai_blindspot}
\end{enumerate}
Though all of these papers make mention of the importance of scrutinizing training data, the emphasis is on model training, deployment, and operationalization.

\subsection{Algorithmic Impact Assessments}
Inspired by environmental impact statements that construction programs must produce, algorithmic impact statements offer an accountability framework for those who operationalize algorithms
\citep{shneiderman_opinion_2016,ainow}. Just as ``the environmental impact statement process combines a focus on core values with a means for the public, outside experts, and policymakers to consider complex social and technical questions'' \citep{ainow}, the algorithmic impact statement advocates for an iterative process of development between agencies, the public, and knowledgeable outside parties. Examples of algorithmic impact assessments include: 
\begin{enumerate}
    \itemsep0em 
    \item The AI Now Institute’s ``Algorithmic Impact Assessments'' \citep{ainow}
    \item The ACM Conference on Fairness, Accountability, and Transparency in Machine Learning's ``Principles for Accountable Algorithms'' and ``Social Impact Statement for Algorithms'' \citep{fatml}
    \item Nick Diakopoulos's ``Algorithmic Accountability Reporting'' \citep{diakopoulos_algorithmic_2014, diakopoulos_2016}
\end{enumerate}
It should be noted that the algorithmic impact assessment has a strong emphasis on policy applications and thus tends to be written with policy makers in mind.

\subsection{Toolkits \& Ethics-based Approaches}

A fourth category of assessment is the algorithmic ``toolkit,'' which does not fall cleanly into the three aforementioned categories. Two examples from this category are summarized below:

\begin{enumerate}
    \itemsep0em 
    \item Microsoft Research’s FairLearn toolkit \citep{bird2020fairlearn}, which could be considered part of the model assessment class, but which I have differentiated because it is a codebase that practitioners can use to audit their own systems from the perspective of fairness. Using the toolkit does not result in a deliverable to be shared with the public but rather helps the practitioner to modify the system itself. 
    \item The Washington State ACLU’s Algorithmic Equality Toolkit \citep{katell_algorithmic_2019}, which is differentiated from the model assessment category because it is intended primarily for activists and community advocates in order to ``promote public understanding of algorithms and artificial intelligence'' and increase accountability and regulation.
\end{enumerate}
Related to the toolkits in this category are ethics-based approaches to tech project design, which offer slightly different but valuable perspectives to the categories previously enumerated. Examples include ``Envisioning Cards'' \citep{friedman_envisioning_2012}, ``Tarot Cards for Tech'' \citep{tarot_cards}, and ``Surveying the Landscape of Ethics-Focused Design Methods'' \citep{chivukula2021surveying}, a survey of 63 such methods.

\section{Selecting Representative Projects for Case Studies}\label{sec:results2}

In order to develop a checklist for machine learning and cultural heritage, it was next necessary to identify a representative selection of existing projects in this emerging body of work in order to serve as case studies. To identify relevant projects, I consulted four reports that survey examples in this space: the Library of Congress Labs team’s summary report for the ML + Libraries summit  \citep{eventsummaryLCLabs}, Ryan Cordell’s report ``Machine Learning + Libraries: A Report on the State of the Field'' \citep{cordell_machine_nodate}, Thomas Padilla’s OCLC report ``Responsible Operations: Data Science, Machine Learning, and AI in Libraries'' \citep{padilla_responsible_2020}, and the ``EuropeanaTech AI in relation to GLAMs'' Task Force’s report \citep{europeanaInterimReport}. In addition, I consulted the digital humanities project registry published by \textit{Reviews in the Digital Humanities} \citep{reviews_in_dh}, and I informally surveyed colleagues to generate candidate projects.

It is important to clarify that the very notion of a ``representative'' subsample of projects is a fraught one, as it is highly subjective and entirely dependent on one's criteria for what constitutes representativity. To address this, I refined dimensions and criteria along which I evaluated each project. These criteria are enumerated below and are inspired by dimensions facets in the reports that I consulted:
    \begin{itemize}
        \itemsep0em 
        \item[--] \textit{Institutional setting} (libraries, archives, museums, academic departments)
        \item[--] \textit{Digitized collection medium} (images, text, video, audio)
        \item[--] \textit{Digitized collection topic}, \textit{including time period, subject matter, geographic location, and language}
        \item[--] \textit{Machine learning methodology employed} (image classification, facial recognition, named entity recognition, etc.)
        \item[--] \textit{Discipline or field of study} (history, computer science, data art, etc.)
        \item[--] \textit{Intended audience} (historians, educators, the public)
        \item[--] \textit{Deliverable form} (paper, visualization, interface, exhibit) 
    \end{itemize}
    
\begin{table}[htbp]
  \centering
  \footnotesize
    \begin{tabular}{ | p{4cm} | p{2.5cm} | p{2.2cm} | p{2.3cm} | p{2.5cm} | }
    \hline
    \textbf{Project} & \textbf{Collection Medium} & \textbf{ML Task} & \textbf{Audience} & \textbf{Deliverable Form} \\ \hline
    \textbf{The Real Face of White Australia} & Document Scans (Gov't Documents) & Facial Recognition & Public & ``Wall of Faces'' Online Interface \\ \hline
    \textbf{Citizen DJ} & Audio (Music, Field Recordings, etc.) & Audio Extraction \& Similarity & Public & Online Hip Hop Sampler \& Exploratory Interface \\ \hline
    \textbf{The Distant Viewing project} & Video (TV Sitcoms) & Facial Recognition \& Image Classification (at frame level) & Scholars \& ML Researchers & The Distant Viewing Toolkit \& Scholarly Output \\
    \hline
    \textbf{The Transkribus platform} & Document Scans (Handwritten \& Typewritten) & OCR \& Handwriting Recognition & Scholars, Librarians \& Archivists, ML Researchers & Online Platform \\ \hline
    \textbf{Newspaper Navigator} & Document Scans (Newspapers) & Visual Content Extraction \& Similarity & Scholars, ML Researchers \& Public & The Newspaper Navigator Dataset \& Search Interface + Scholarly Output \\ \hline
    \end{tabular}
    \caption{A table categorizing the selected projects as case studies for the ``Collections as ML Data'' checklist developed in this paper.}\label{tab:projects}
\end{table}

This process yielded five representative projects: 
\begin{enumerate}
    \itemsep0em 
    \item \textbf{\href{http://www.realfaceofwhiteaustralia.net/}{The Real Face of White Australia}}, a project created by Kate Bagnall and Tim Sherratt at the University of Tasmania and the University of Canberra, respectively \citep{sherratt_tim_2019_3579530}. The project utilizes facial recognition to uncover photographs of non-white Australians as preserved within the National Archives of Australia in order to ``explore the records of the White Australia Policy through the faces of those people.''
    \item \textbf{\href{https://citizen-dj.labs.loc.gov/}{Citizen DJ}}, a project by Brian Foo, an Innovator in Residence at the Library of Congress and a Data Visualization Artist at the American Museum of Natural History \citep{citizendj}. Citizen DJ uses machine learning to extract and sort audio samples from the Library of Congress's collections and allows the American public to explore the collections by remixing the samples using a hip-hop sampler interface.
    \item \textbf{\href{https://readcoop.eu/transkribus/}{The Transkribus platform}}, initially ``developed by the University of Innsbruck in cooperation with leading research groups from all over Europe as part of the Horizon 2020 EU research project READ'' \citep{transkribus, transkribus_website}. Transkribus empowers users to train their own OCR models in an interactive machine learning fashion using iterative training with custom sets of typewritten and handwritten documents.
    \item \textbf{\href{https://www.distantviewing.org/}{The Distant Viewing project}} by Lauren Tilton and Taylor Arnold at the University of Richmond \citep{arnold_visual_2019}. The Distant Viewing project utilizes facial recognition and image classification to study sitcoms such as ``Bewitched'' and ``I Dream of Genie'' through the lenses of media studies and the digital humanities.
    \item \textbf{\href{https://news-navigator.labs.loc.gov/search}{Newspaper Navigator}} \citep{lee_compounded_2020,nn_demo,nn_dataset}, my own project developed in conjunction with the Library of Congress, which utilizes object detection to extract visual content from 16 million historic newspaper pages and reimagines exploratory search by empowering users to train their own interactive machine learners to retrieve images according to user-defined facets. I selected my own project primarily because I could reflect on the subtleties involved as a primary stakeholder.
\end{enumerate}

If a central goal of the ``Collections as ML Data'' checklist is to aid researchers and practitioners in this space, one condition for the success of the checklist is applicability across a range of projects. In Table 1, I compare the selected projects according to the criteria articulated earlier in this section. The range of the selected projects' collection media, machine learning tasks, audiences, and deliverable forms not only speak to the diverse nature of projects within machine learning and cultural heritage but also collectively serve as an important test surrounding the ``Collections as ML Data'' checklist's relevance. Iteratively refining the checklist by applying it to these projects as described in the section ``\nameref{sec:methodology}'' served an important method for conceptualizing the responsible practices detailed and improving the comprehensiveness of the checklist's questions and suggestions. In the section ``\nameref{sec:application},'' I provide vignettes describing the application of the refined ``Collections as ML Data'' checklist to each project, revealing the ways in which each project tests the checklist in a different manner.

\section{The ``Collections as ML Data'' Checklist: An Overview}\label{sec:checklist}

In this section, I provide an overview of the ``Collection as ML Data'' checklist. An outline of the checklist can be found below. The overview in this section is structured around the four central components of the checklist: the cultural heritage collection as data; the machine learning model; organizational considerations; and copyright, transparency, documentation, maintenance, and privacy.  The full ``Collections as ML Data'' checklist is enumerated in the Appendix. The checklist is partitioned into four components; each checklist question that is directly inspired by related work is accompanied by corresponding citations. In this section, I elaborate each of the four components. In the next section, I provide use cases of the checklist.

\small{
\setlength{\multicolsep}{6.0pt plus 2.0pt minus 1.5pt}
\begin{tcolorbox}[sharp corners, colback=green!30, colframe=green!80!blue, title=An Outline of the ``Collections as ML Data'' Checklist]
\begin{enumerate}
    \item \textbf{The Cultural Heritage Collection as Data} 
    \begin{multicols}{2}    
    \begin{enumerate}
        \itemsep0em 
        \item Dataset Composition
        \item  Collecting Process \& Curation Rationale
        \item Digitization Pipeline
        \item Data Provenance
        \item Crowd Labor
        \item Additional Modification
    \end{enumerate}
    \end{multicols}
    \item \textbf{The Machine Learning Model} 
    \begin{multicols}{2}
    \begin{enumerate}
        \itemsep0em 
        \item Overview
        \item Training / Finetuning
        \item Evaluation
        \item Deployment
        \item Release
        \item Environmental Impact
    \end{enumerate}
    \end{multicols}
    \item \textbf{Organizational Considerations} 
    \begin{multicols}{2}
    \begin{enumerate}
        \itemsep0em 
        \item Stakeholders
        \item Use of Machine Learning
        \item Organizational Context
        \item Project Deployment \& Launch
    \end{enumerate}
    \end{multicols}
    \item \textbf{Copyright, Transparency, Documentation, Maintenance, and Privacy}
    \begin{multicols}{2}
    \begin{enumerate}
        \itemsep0em 
        \item Copyright
        \item Transparency \& Re-Use
        \item Documentation
        \item Maintenance
        \item Privacy
    \end{enumerate}
    \end{multicols}
\end{enumerate}
\end{tcolorbox}
}
\normalsize 

\subsection{The Cultural Heritage Collection as Data}\label{sec:overview_1}

In this component of the ``Collections as ML Data'' checklist, practitioners are encouraged to interrogate and reflect on the cultural heritage collection(s) being utilized in the project as data. In drawing from the dataset assessments described in the section ``\nameref{sec:data},'' this component emphasizes a clear understanding of the dataset's composition and provenance, including who is represented in the dataset, how the data was collected, and beyond. In accordance with the nuances required in the context of cultural heritage, this component also draws from best practices in cultural heritage in order to encourage practitioners to excavate curation, digitization, and any crowd labor utilized in augmenting the dataset. By articulating the contours of the cultural heritage data in question, practitioners engage with the sociotechnical implications of treating cultural heritage collections as data, producing documentation that foregrounds these considerations for the project's audience. 

\subsection{The Machine Learning Model}\label{sec:overview_2}

In this component of the checklist, practitioners are asked to engage with the subtleties of the machine learning model(s) utilized. The questions in this section draw from the model assessments described in the section ``\nameref{sec:model}'' in order to address canonical questions surrounding a model's documentation, including its training, evaluation, deployment, and release. This component is tailored to the specifics surrounding cultural heritage collections, including whether the machine learning model has been utilized to make a single pass over a collection or will be continuously deployed, and whether the model has applications outside of cultural heritage. Moreover, this component asks the practitioner to consider the environmental impact of training and deploying the model. Collectively, the questions in this section emphasize both scientific and ethical best practices.

\subsection{Organizational Considerations}\label{sec:overview_3}

In this component of the checklist, practitioners are encouraged to consider the broader organizational considerations of the project, including documenting the stakeholders and organizational context. Because projects may involve stakeholders with different fluencies and experience levels with both machine learning and cultural heritage, this component emphasizes considerations surrounding the subtleties that emerge in this context: do stakeholders have access to gain expertise in these domains? Can this project be utilized to build data fluency at the organization? Moreover, stakeholders are asked to consider a fundamental question that is often overlooked in the machine learning literature: is it necessary to use machine learning in this context, and if so, why? Lastly, the stakeholders are asked to consider the target audiences of the project. Here, stakeholders are asked to return after a project's launch in order to reflect on which audiences were reached and what feedback was received. More generally, this component encourages stakeholders to engage critically with the organizational complexities introduced at the intersection of machine learning and cultural heritage.

\subsection{Copyright, Transparency, Documentation, Maintenance, and Privacy}\label{sec:overview_4}

The fourth and final component of the ``Collections as ML Data'' checklist concerns five project components that are essential to a successfully launched project: copyright, transparency, documentation, maintenance, and privacy. Given the subtleties raised by copyright in the context of cultural heritage, stakeholders are asked to engage with the ways in which copyright impacts the project's scope and deliverables. Stakeholders are also encouraged to walk through the project's efforts toward transparency, including support for outside audits, as well as the availability of code documentation for reproducibility and reuse. Because cultural heritage institutions tend to have stringent considerations surrounding user privacy, stakeholders are asked to report what data on project visitors will be collected and whether consent will be asked for. Lastly, stakeholders are asked to address plans for maintenance after launch. This final component of the ``Collections as ML Data'' checklist emphasizes a holistic approach to project development that is often overlooked when considering only the machine learning elements in isolation.

\section{Applying the ``Collections as ML Data'' Checklist}\label{sec:application}

In order to justify four components of the ``Collections as ML Data'' checklist and provide concrete case studies, I have included vignettes from each of the projects articulated in Table 1. These vignettes illustrate how adopters of the ``Collections as ML Data'' checklist might begin utilizing the guiding questions.

\subsection{Case Studies 1:  The Cultural Heritage Collection as Data}

Building on the section ``\nameref{sec:overview_1},'' the following vignettes demonstrate the subtleties raised by treating cultural heritage collections as data for machine learning projects, whether for training a machine learning model or for processing with one.

    \begin{enumerate}
        \item \textit{Dataset Composition}: Though the questions in this section are inspired by machine learning-oriented checklists \citep{data_statements_NLP, gebru2020datasheets, holland2018dataset}, understanding the composition of a cultural heritage dataset is crucial to \textit{any} cultural heritage project and research. Machine learning only redoubles the considerations that must be made. ``Citizen DJ'' is an exemplary project from the perspective of how a collection's composition must be taken into consideration. To help guide visitors using the project's hip-hop sampler, Brian Foo has created an ethics guide that is available on the project's main site \citep{citizendjguide}. The guide walks a visitor through the process of considering a dataset's composition in order to properly address attribution, compensation, and cultural and historical contexts. 
        \item  \textit{Collecting Process \& Curation Rationale}: Many cultural heritage collections have decades-long, complex histories surrounding their creation and curation. First, let us consider a collection's creation. In the case of ``The Real Face of White Australia,'' the government documents in consideration were originally produced under the Australian Immigration Restriction Act, namely, certificates granting exemption to the Dictation Test \citep{sherratt_tim_2019_3579530}. The project's goal of recovering the people marginalized by these policies can only be understood when considering the documentation's role within the oppressive system. The project is notable for how the project leaders pay close attention to the origins of a collection that documents a difficult history. Next, let us turn to curation. In the case of \textit{Chronicling America}, the newspaper corpus on which Newspaper Navigator is built, the selection process for including a newspaper title within the collection is a highly nuanced process, dependent on criteria enumerated by the Division of Preservation and Access at the National Endowment for the Humanities, as well as state-level contributors \citep{lee_2020}. Understanding these curation decisions are essential to properly contextualizing the abundances and lacunae of representation within the corpus.
        \item \textit{Digitization Pipeline}: The digitization pipeline can have a profound impact on a machine learning model's predictions, as evidenced by the Newspaper Navigator dataset \citep{nn_dataset, lee_compounded_2020}. Lyneise Williams has documented how the distortive effects of the microfilming process can lead to erasure of people of color by saturating darker skin tones \citep{williams_CAS}. This phenomenon is present within \textit{Chronicling America} and digital cultural heritage collections writ large. In particular, the Newspaper Navigator data archaeology demonstrates how embeddings generated by a ResNet model pre-trained on ImageNet fail to retrieve the same photograph of W.E.B. Du Bois among four different digitized \textit{Chronicling America} newspaper pages due to precisely this effect of microfilming distortion. Documenting the digitization process is thus crucial to understanding how a machine learning model processes a cultural heritage dataset \citep{lee_compounded_2020}.
        \item \textit{Data Provenance}: Though much of a cultural heritage dataset's provenance will have been articulated in the previous two sections (``Collecting Process \& Curation Rationale'' and ``Digitization Pipeline''), there might be additional salient details regarding the genealogy of the data. Consider, for example, the Library of Congress's ``American English Dialect Recordings: The Center for Applied Linguistics Collection'' included within Citizen DJ. The recordings in the collection were originally obtained by over 200 collectors. In 1983, the Center for Applied Linguistics obtained these recordings from the collectors in order to improve access, as funded by a grant from the National Endowment for the Humanities. In 1986, the Center for Appplied Linguistics donated 405 audio recordings to the Library of Congress, 350 of which have been made available online (the remaining 55 withheld due to copyright considerations). The complex provenance of collections must therefore be considered when assessing a digital collection's contours.
        \item \textit{Crowd Labor}: Many cultural heritage institutions are pursuing volunteer crowdsourcing initiatives to engage the public with their collections. In the case of Newspaper Navigator, the bounding boxes utilized as training data for the visual content recognition model was derived from \textit{Beyond Words}, a crowdsourcing initiative launched in 2017 by LC Labs that asked volunteers to identify photographs, illustrations, maps, comics, and editorial cartoons on World War 1-era newspaper pages in \textit{Chronicling America}. Because volunteers find the experience an enriching and educational opportunity, the experience has something to offer to both the Library of Congress and the American public. This initiative stands in sharp contrast to the utilization of outsourced, contracted laborers to improve metadata with datasets, an approach common among machine learning datasets in the machine learning community. Crowd workers such as Mechanical Turk workers are paid extremely low hourly wages \citep{hara2017datadriven}. This checklist section is thus provided to encourage project stakeholders to consider the project's relationship to labor and to improve transparency to the project's audience.
        \item \textit{Additional Modification}: Because a cultural heritage dataset might require additional modifications in order to compute against it, the question in this section is intended to provide the project stakeholders with an opportunity to document any changes made.
    \end{enumerate}
    
Collectively, these vignettes provided in response to the ``Collections as ML Data'' checklist draw out nuances that would not necessarily be surfaced by responding to a canonical dataset assessment from the machine learning literature. As demonstrated by these examples, data provenance, curation, and processing via digitization pipelines represent nontrivial steps within cultural heritage and must be foregrounded in a project's documentation -- as the ``Collections as ML Data'' checklist encourages.

\subsection{Case Studies 2: The Machine Learning Model}

The following vignettes elaborate on the section ``\nameref{sec:overview_2}'' in order to justify the importance of documenting the machine learning model(s) employed across each facet of this ``Collections as ML Data'' checklist component. All five case studies are highlighted with the goal of foregrounding the diversity of responses that the ``Collections as ML Data'' checklist elicits surrounding the machine learning components of projects.

\begin{enumerate}
    \item \textit{Model Details}: The questions in this section are intended to provide an overview of the machine learning model(s) being utilized, including: \textit{What architectures are being used? What tasks are they being used for?} These questions are standard within the machine learning literature, and all of the case studies address the questions in this section within their documentation.  
    \item \textit{Training / Finetuning}: Projects at the intersection of machine learning and cultural heritage often differ from machine learning research, in that the former do not always emphasize methodological advances in machine learning. As a result, these projects often utilize off-the-shelf algorithms or models, as is the case with Citizen DJ \citep{citizendj}, ``The Real Face of White Australia'' \citep{sherratt_tim_2019_3579530}, and the Distant Viewing project \citep{arnold_visual_2019}. In the case of Newspaper Navigator, a pre-trained object detection model was finetuned for the specific task of identifying visual content on newspaper pages \citep{nn_dataset}. In the case of Transkribus, end-users are empowered to train their own OCR and handwriting recognition models on datasets that they themselves have curated. All of these projects elaborate on the considerations surrounding the specific approach chosen.
    \item \textit{Evaluation}: A model's evaluation is already standard practice in machine learning, and this section contains questions that are already operationalized by the field writ large. The main provocations in this section are intended to encourage project stakeholders to consider auditing their systems for fairness and utilizing tools for generating explanations for predictions if necessary.
    \item \textit{Deployment}: In the context of cultural heritage projects, the deployment of a machine learning model can take many forms: from the one-off utilization of a machine learning model to produce metadata or pre-process data (as utilized by all projects considered in this section) to the deployment of machine learning models that users can continuously train and use to transcribe scan documents (as is the case with Transkribus \citep{transkribus_website}) or re-rank image search results (as is the case with the Newspaper Navigator search application \citep{nn_demo}). These varied uses motivate the guiding questions in this section regarding deployment.
    \item \textit{Release}: This section serves to elicit elaborations surrounding the model's release, from availability to extensibility. Transkribus's model for public release is particularly notable: Transkribus users have made 58 trained models publicly available to date \citep{transkribus_model_website}. These models are contributed to the central repository along with corresponding documentation, including who trained each model, what training dataset was utilized, benchmark evaluation scores, and the types of script that the model can process. 
    The Newspaper Navigator finetuned visual content recognition model weights have been incorporated into LayoutParser, a pip-installable Python package for document layout analysis \citep{shen2020layoutparser}, presenting yet another option for release and re-use.
    \item \textit{Environmental Impact}: It has been well-documented that the training and deployment of machine learning models can have significant carbon footprints, commensurate with vehicular emissions \citep{strubell_2019, schwartz2019green}. In the case of the Newspaper Navigator dataset construction, over 5 years of wall clock computing time was required to produce the dataset \citep{nn_dataset}. The Newspaper Navigator data archaeology documents the emissions produced by the project from training through deployment \citep{lee_compounded_2020}. In total, this amounted to approximately 226 kg of CO$_2$ emissions, equivalent to a single cross-country flight. As machine learning projects continue to become more computationally intensive, considering the carbon emissions of a machine learning model and pursuing alternative, less resource-intensive approaches are responsibilities for which all project stakeholders must be held accountable.
\end{enumerate}

Given the range of uses of machine learning demonstrated by the case studies and the different responses that the ``Collections as ML Data'' checklist could elicit, it is evident that proper documentation using the checklist can help guide stakeholders, project audiences, and external auditors regarding the specifics of how machine learning has been utilized.

\subsection{Case Studies 3: Organizational Considerations}

Following the section ``\nameref{sec:overview_3},'' these examples serve to demonstrate the value of documenting a project's organizational considerations using the ``Collections as ML Data'' checklist. As demonstrated below, this component surfaces organizational context that might otherwise not be documented.

\begin{enumerate}
    \item \textit{Stakeholders}: The questions in this section are intended to capture the intricacies surrounding project stakeholders, from considering stakeholder backgrounds to reflecting on whether all relevant stakeholder groups have been included. The ``Real Face of White Australia'' details these considerations surrounding stakeholder backgrounds in their chapter ``The People Inside'' \citep{sherratt_tim_2019_3579530}, namely, the experience of the historian learning how to utilize machine learning within the context of the project. Of note is the project's considerations surrounding the decision to present individuals' photographs even though the individuals cannot consent, as the photos are from over a century ago. The explicit surfacing of this question surrounding the consultation of stakeholder groups foregrounds the challenges that a project in this space must confront.  
    \item \textit{Use of Machine Learning}: Within the cultural heritage community, special  considerations are taken when applying computational methodologies \citep{fogu_probing_2016}. Within this context, it is imperative to consider why machine learning must be applied. Is the motivation driven by real need or by organizational pressures? In the case of Newspaper Navigator, the data archaeology \citep{lee_compounded_2020} motivates the utilization of machine learning: not only to improve access at scale but also to re-imagine the search affordances by empowering users to train their own machine learners to retrieve relevant visual content. 
    \item \textit{Organizational Context}: The questions in this section are motivated by the recommendations and guiding questions in the reports detailing best practices for machine learning and cultural heritage enumerated in the section ``\nameref{sec:related_work}.'' The intent is to ask stakeholders to reflect on how the project can serve the broader institution or organization by improving data fluency and training surrounding both machine learning and cultural heritage. Doing so can have significant longitudinal effects toward the proper operationalization of machine learning in cultural heritage institutions and proper ethical considerations surrounding cultural heritage at organizations specializing in machine learning. In the case of Newspaper Navigator, the project's organizational context is documented in an article in \textit{EuropeanaTech Insight} \citep{europeanatechinsight}, detailing how the project fits into the Library of Congress's digital strategy \citep{lcdigitalstrategy}.
    \item \textit{Project Deployment \& Launch}: The questions in this section are split into two categories: pre-launch and post-launch. The intent is to ask project stakeholders to reflect honestly about the goals surrounding the project's launch and the successes and failures relative to these goals.  
\end{enumerate}

As revealed through these case studies, additional nuances emerge for projects in this space surrounding fluency in both machine learning and cultural heritage at the stakeholder and organizational level. With the ``Collections as ML Data'' checklist, project stakeholders are encouraged to surface these subtleties and critically engage with their motivations for utilizing machine learning in the project as well.

\subsection{Case Studies 4: Copyright, Transparency, Documentation, Maintenance, and Privacy}

\noindent
Building on ``\nameref{sec:overview_4},'' the following vignettes demonstrate the importance of considering a project's deliverables beyond the typical framings of data assessments and model assessments. 

\begin{enumerate}
    \item \textit{Copyright}: Copyright is an essential consideration of any ML project involving cultural heritage collections, from the collections themselves to the machine learning models, code, and final deliverables. Citizen DJ illustrates the subtleties introduced by copyright. Brian Foo has created a copyright checklist for users to assess how samples created within Citizen DJ can be re-used \citep{citizendjguide}. Moreover, all of the project's code is open source and placed in the public domain for unrestricted re-use, meaning that other cultural heritage practitioners as well as members of the general public can utilize the code for their own projects. The questions in this section of the checklist are included in order to have project stakeholders document the subtleties introduced by copyright in its many manifestations.
    \item \textit{Transparency \& Re-use}: Making a project's code and deliverables transparent and extensible for re-use is a valuable contribution to the ``Collections as Data'' community. The Distant Viewing Toolkit is exemplary in this regard \citep{Arnold2020}. The toolkit not only powers the Distant Viewing Project but also makes it possible for other digital humanities and media studies scholars to utilize the toolkit for their own research projects. Both Citizen DJ and Newspaper Navigator have made all code for the project available as well, enabling external auditors to evaluate the systems.
    \item \textit{Documentation}: Documentation is essential to promoting transparency and facilitating re-use. ``The Real Faces of White Australia,'' Transkribus, Newspaper Navigator, Citizen DJ, and the Distant Viewing project all have code documentation in the form of publicly-available GitHub repositories. In addition, ``The Real Faces of White Australia,'' Newspaper Navigator, Citizen DJ, and the Distant Viewing projects have documented the methodologies employed and the socio-technical contexts for the projects \citep{citizendjguide, arnold_visual_2019, sherratt_tim_2019_3579530, nn_dataset, lee_compounded_2020}. This documentation makes it possible for the project's audiences and potential auditors to understand the context and considerations taken by the project stakeholders. Moreover, code documentation facilitates open re-use.
    \item \textit{Maintenance}: Because the deliverables of many projects at the intersection of machine learning and cultural heritage are digital artifacts, providing clear documentation on expected maintenance and upkeep is important for the project's audience. Citizen DJ details the project's expected lifecycle on its `About' page, thereby helping site visitors to set realistic expectations surrounding sustained usage of the tool.
    \item \textit{Privacy}: For any project deployed on the internet, it is important to provide users with an understanding of what data are being collected and ask for consent if necessary. For cultural heritage projects, this is particularly sensitive, given the long history of libraries' dedication to preserving patron privacy. For both Citizen DJ and Newspaper Navigator, no personally-identifiable information is collected on any site visitor, in compliance with the Library of Congress's privacy policy. 
\end{enumerate}

As revealed through these case studies, the ``Collections as ML Data'' checklist encourages stakeholders to document all aspects of a project's deliverables, especially through the lenses of transparency, sustainability, and privacy. By centralizing this documentation, the checklist encourages practitioners, as well as the broader community, to foreground these considerations.

\section{Discussion \& Future Work}

In this paper, I have introduced the ``Collections as ML Data'' checklist, a detailed set of guiding questions for projects that utilize machine learning in the context of cultural heritage collections. The checklist serves as a conceptual model for ethically responsible decisions in the context of applying machine learning to cultural heritage collections. When completed by stakeholders, the checklist answers can be published along with the deliverables of the project in order to increase transparency and foreground responsible practices. Such a conceptual model is particularly important in this space, where ethical failures of machine learning projects can be redoubled due to the sensitivities required with cultural heritage collections in particular. 

In the section ``\nameref{sec:related_work},'' I began the paper by contextualizing the ``Collections as ML Data'' checklist within two bodies of emerging work: the machine learning movement to produce checklists, toolkits, and algorithmic impact statements, as well as the formation of best practices and state-of-the-field reports from the cultural heritage community for utilizing machine learning. In the section ``\nameref{sec:methodology},'' I then described my methodology in producing the ``Collections as ML Data'' checklist. This process included detailing a taxonomy of existing machine learning guidelines in order to inform the contributions of the ``Collections as ML Data'' checklist, as well as to provide cultural heritage practitioners with a guide to existing work within the machine learning literature. I also documented the process by which I selected five projects to serve as case studies for developing and evaluating the checklist. I then provided an overview of the ``Collections as ML Data'' checklist itself and provided concrete examples of applying components of the checklist to the five case studies in order to demonstrate the value of the checklist. The remaining sections serve to reflect on the ``Collections as ML Data'' checklist's use in practice, how the checklist fits into the field of machine learning more broadly, and future work.

\subsection{Applying the ``Collections as ML Data'' Checklist in Practice}

The ``Collections as ML Data'' checklist is intended to be utilized as a resource throughout all stages of a machine learning project with cultural heritage data, from the initial steps of identifying a digital collection to the final steps of publishing the project's deliverables. Per the recommendations of \citep{fatml}, consultation and engagement with the checklist questions would ideally take place during the design phase, pre-launch, and post-launch in order to allow for the checklist's considerations to be operationalized and the checklist responses to be filled out and refined during the project's development. Moreover, a completed version can be published along with the project's deliverables in order to promote transparency, as well as communicate decisions made and shortcomings faced during the project's development with both the project's audience and potential auditors.

\subsection{Reflecting on Machine Learning and Cultural Heritage}

Though the ``Collections as ML Data'' checklist has been formulated for projects involving humanists and cultural heritage practitioners as stakeholders, it should be noted that cultural heritage is, in a sense, the substrate of machine learning. From the text on the web to the photos on Flickr, our collective cultural heritage is utilized ubiquitously by machine learning researchers and product teams across the world \citep{mahajan_exploring_2018, brown2020language}. In this regard, the specific considerations surrounding cultural heritage that are offered in the ``Collections as ML Data'' checklist can be interpreted much more broadly. 

\subsection{Future Work}

Though the ``Collections as ML Data'' checklist has been subject to many iterations, it is nonetheless not comprehensive. Based on feedback from readers and practitioners who utilize the checklist, I hope to refine it even further and maintain a list of additional considerations for stakeholders to consider. Of course, no checklist can be entirely comprehensive, and such a resource is valuable but certainly not sufficient: just because a checklist has been utilized does not mean the project should not be interrogated further or documented more extensively. I will conclude by paraphrasing Ryan Cordell's first guiding question in his ``25 Questions for Structuring an ML Project'': what is \textit{missing} from the ``Collections as ML Data'' checklist \citep{cordell_machine_nodate}? 

\section{Acknowledgments}

This research is based upon work supported by the National Science Foundation Graduate Research Fellowship under Grant DGE-1762114. I would like to thank Professor Nic Weber, Professor Katharina Reinecke, and the CSE 599 ``Computer Ethics'' class at the University of Washington, as well as Professor Marti Hearst at the University of California, Berkeley, for their invaluable feedback, advice, and guidance throughout the development of this paper. I would also like to thank my Ph.D. advisor, Professor Daniel Weld, at the University of Washington and the LC Labs team at the Library of Congress for formative discussions surrounding this work.

\section{Appendix: The Full ``Collections as ML Data'' Checklist}\label{sec:full_checklist}

\subsection{The Cultural Heritage Collection as Data}\label{sec:checklist_dataset}

\textit{Here, a distinction is drawn between the cultural heritage collection being studied and the training dataset being utilized for the machine learning model. For example, a project might utilize a pre-trained model to generate embeddings for a photo collection. In this section, we consider the cultural heritage collection itself; in the section ``\nameref{sec:model_checklist},'' we consider the machine learning model's training data.}

\begin{enumerate}
    \itemsep0em 
    \item \textbf{Dataset Composition}
    \begin{enumerate}
        \itemsep0em 
        \item Who or what are depicted in the dataset? \citep{gebru2020datasheets}
        \item If the dataset depicts people, are any specific subgroups of people represented? Are any specific individuals personally identifiable? \citep{gebru2020datasheets}
        \item If the dataset depicts people, are any individuals still living? Does this project comply with privacy laws in countries where it will be shared?
        \item What medium is the dataset? (image, video, text, web archive, etc.)
        \item How large is the dataset, both in cardinality and in disk storage?
        \item What metadata is available for the dataset items? \citep{holland2018dataset}
        \item Does copyright impact this dataset? If so, how? \citep{padilla_2018, cordell_machine_nodate, eventsummaryLCLabs, gebru2020datasheets}
        \item Does this dataset pertain to a difficult history? If so, what extra precautions are being taken?
    \end{enumerate}
    \item \textbf{Collecting Process \& Curation Rationale} (\textit{language borrowed from} \citep{data_statements_NLP})
    \begin{enumerate}
        \item Who curated the cultural heritage collection from which this dataset is derived?
        \item What organization or institution was the collection created for?
        \item What funding was utilized (if known)?
        \item What collection process was utilized? \citep{data_statements_NLP}
        \item When was the collection assembled? (i.e., when were the photographs taken or ethnographies recorded?) 
        \item What instruments were utilized to create the collection? (i.e., a recording device, camera, etc.)
        \item If people are included, did individuals consent at the time of collection?
        \item What were the decision-making processes behind the collection’s curation? \citep{data_statements_NLP}
        \item What is unknown about the collection process \& curation rationale?
    \end{enumerate}
    \item \textbf{Digitization Pipeline} (only applicable if the dataset is a digitized version of a physical collection)
    \begin{enumerate}
        \item Who selected what was digitized?
        \item What organization or institution oversaw the digitization?
        \item What funding was utilized?
        \item What criteria were utilized for determining what was digitized? \citep{cordell_machine_nodate}
        \item What were the steps in the digitization pipeline? (For example, in the case of photos, what scanners were used to digitize the documents? In the case of documents, what OCR engines were utilized?)
        \item What metadata was algorithmically produced?
    \end{enumerate}
    \item \textbf{Data Provenance}
    \begin{enumerate}
        \item What is the provenance of the dataset, from collection through digitization? \citep{data_statements_NLP,holland2018dataset,fatml}
        \item Is any part of the provenance unknown?
    \end{enumerate}
    \item \textbf{Crowd Labor}
    \begin{enumerate}
        \item Have volunteers or crowd workers added metadata to the dataset? \citep{eventsummaryLCLabs, cordell_machine_nodate, padilla_2018}
        \item If so, how were they recruited and compensated?
        \item If so, what metadata did they produce? (i.e., transcriptions, annotations, etc.)
    \end{enumerate}
    \item \textbf{Additional Modification}
    \begin{enumerate}
        \item Were any additional steps taken after collection curation and digitization in order to produce the dataset in question? (i.e., Were any items removed? Were any additional metadata added? etc.)
    \end{enumerate}
\end{enumerate}

\subsection{The Machine Learning Model}\label{sec:model_checklist}
\textit{Note: if multiple machine learning models were utilized in the project, this step should be completed for each model.}

\begin{enumerate}
    \item \textbf{Overview}
    \begin{enumerate}
        \item What model architecture has been utilized? \citep{mitchell_model_2019}
        \item What is the task that the model is being deployed to perform?
        \item Who trained, finetuned, and/or deployed this model? \citep{mitchell_model_2019}
        \item Across what organizations or institutions did this training, finetuning, and/or deployment take place? \citep{mitchell_model_2019}
        \item What funding was utilized? \citep{gebru2020datasheets}
    \end{enumerate}
    \item \textbf{Training / Finetuning}
    \begin{enumerate}
        \item Was the model trained from scratch?
        \item If so, what data was used to train the model? \citep{mitchell_model_2019}
        \item If not, was a pre-trained model utilized? Where can more information on the pre-trained model be found?\citep{mitchell_model_2019}
        \item Was the pre-trained model finetuned? If so, what data was utilized for finetuning?
        \item If training or finetuning was performed, what computational resources were utilized?
    \end{enumerate}
    \item \textbf{Evaluation}
    \begin{enumerate}
        \item How was the model’s performance evaluated? \citep{mitchell_model_2019}
        \item What data was used for evaluation? \citep{mitchell_model_2019, arnold2019factsheets}
        \item If the model involves data pertaining to people, has the model been audited for fairness and bias using tools such as FairLearn? \citep{arnold2019factsheets, bird2020fairlearn, ainow, fatml, madaio_co-designing_2020, eventsummaryLCLabs}
        \item Have any tools been utilized to generate explanations for predictions (i.e., LIME \citep{lime}, SHAP \citep{lundberg_shap}, TCAV \citep{tcav}) and modify the model in response? \citep{arnold2019factsheets, cordell_machine_nodate,padilla_responsible_2020, fatml,ribeiro_beyond_2020}
    \end{enumerate}
    \item \textbf{Deployment}
    \begin{enumerate}
        \item How was the model deployed? Was it used to make a single pass over the cultural heritage dataset in question, or will it be continuously deployed?
        \item What computational resources were utilized for deployment?
        \item Are the metadata generated by the machine learning model (embeddings, classifications, etc.) available as project deliverables?
    \end{enumerate}
    \item \textbf{Release}
    \begin{enumerate}
        \item Has the resulting model been made available for download? (\textit{if no, the following questions can be skipped})
        \item What license has been provided? \citep{mitchell_model_2019}
        \item Who are the primary intended users, and what are the intended use cases? \citep{mitchell_model_2019}
        \item Does this model have applicability outside of cultural heritage collections?
        \item What are ways that this model could be misused, either intentionally or unintentionally? \citep{msr_2020,mitchell_model_2019}
    \end{enumerate}
    \item \textbf{Environmental Impact}
    \begin{enumerate}
        \item What were the carbon emissions produced by training, finetuning, and/or deploying this model? \citep{cordell_machine_nodate, strubell_2019, lacoste2019quantifying}
        \item How does the environmental impact of this model compare to that of other components of the project, such as a collection's digitization or stakeholders' flights to relevant conferences?
    \end{enumerate}
\end{enumerate}

\subsection{Organizational Considerations}\label{sec:organizational_checklist}

\begin{enumerate}
    \item \textbf{Stakeholders}
    \begin{enumerate}
        \item What stakeholder groups are involved in this project? \citep{cordell_machine_nodate}
        \item What is each project member’s familiarity with machine learning? \citep{cordell_machine_nodate, eventsummaryLCLabs}
        \item What is each project member's familarity with cultural heritage collections as data?
        \item Has the project notified and sought input from all potentially relevant stakeholder groups, such as those included within the cultural heritage dataset itself? \citep{ainow, msr_2020}
        \item Do groups affected by the project, such as individuals and communities directly represented within the cultural heritage dataset, have an avenue for contacting project staff and seeking recourse? If so, whom should they contact? If not, why not? \citep{ainow, mitchell_model_2019, fatml}
    \end{enumerate}
    \item \textbf{Use of Machine Learning}
    \begin{enumerate}
        \item Was it necessary to use machine learning for this project?
        \item If so, why?
        \item If not, why was machine learning still utilized?
        \item What are potential critiques of applying machine learning in this context?
    \end{enumerate}
    \item \textbf{Organizational Context}
    \begin{enumerate}
        \item Can this project be used to build data fluency within the organization or institution? \citep{padilla_responsible_2020}
        \item Do there exist programs or paths for training staff affiliated with the project to develop machine learning skillsets? \citep{cordell_machine_nodate,padilla_responsible_2020}
        \item Do there exist programs or paths for training staff affiliated with the project to develop fluency with cultural heritage collections?
    \end{enumerate}
    \item \textbf{Project Deployment \& Launch}
    \begin{enumerate}
        \item Who is the target audience of this project? \citep{msr_2020}
        \item How does the target audience align with the audiences that the institution or organization is hoping to engage?
        \item If the target audience of the project is the public, does it make an attempt to educate the public regarding the machine learning approaches employed?
        \item \textit{Did the project launch reach the intended audience?*}
        \item \textit{Has the project received feedback from stakeholders, including the audience? If so, what feedback has been received?*}
        \item \textit{Has the launch of the project resulted in any changes to the project?*}
    \end{enumerate}
    (* = \textit{to be completed post-launch})
\end{enumerate}

\subsection{Copyright, Transparency, Documentation, Maintenance, and Privacy}\label{sec:last_checklist}

\begin{enumerate}
    \item \textbf{Copyright}
    \begin{enumerate}
        \item Building on question 1.1.c, does copyright impact the dataset, model, code, or deliverables for the project? \citep{gebru2020datasheets, mitchell_model_2019, cordell_machine_nodate, padilla_2018, eventsummaryLCLabs} 
        \item If they are made available, what licenses have been chosen? 
        \item If they are proprietary, how does this impact re-use? 
    \end{enumerate}
    \item \textbf{Transparency \& Re-use}
    \begin{enumerate}
        \item Can the project be audited by outsiders? If so, is there funding available to support outside audits? \citep{ainow, mitchell_model_2019}
        \item Is the code created for the project extensible for other cultural heritage practitioners? \citep{padilla_responsible_2020} \item If so, does the project provide any tutorials or toolkits for re-use?
    \end{enumerate}
    \item \textbf{Documentation}
    \begin{enumerate}
        \item Does the project have documentation? \citep{katell_algorithmic_2019}
        \item If so, is the documentation interpretable by the project's audience?
        \item Is the project reproducible to an outside researcher, given the documentation available?
    \end{enumerate}
    \item \textbf{Privacy}
    \begin{enumerate}
        \item If the project is hosted online, are data on visitors collected? If so, what kinds of user data are collected?  \citep{cordell_machine_nodate}
        \item Is visitor consent gained before gathering online data? \citep{cordell_machine_nodate}
    \end{enumerate}
    \item \textbf{Maintenance}
    \begin{enumerate}
        \item Will the project and code be maintained? \citep{gebru2020datasheets}
        \item If so, how frequently, and who will be responsible for maintaining it?
    \end{enumerate}
\end{enumerate}

\theendnotes

\bibliography{bibliography}

\end{document}